\documentclass[10pt,twocolumn,letterpaper]{article}

\usepackage{wacv}              

\usepackage{graphicx}
\usepackage{amsmath}
\usepackage{amssymb}
\usepackage{booktabs}
\usepackage{mathtools}
\usepackage{multirow}
\usepackage {arydshln}
\usepackage {comment}
\usepackage[dvipsnames]{xcolor}

\definecolor{wacvblue}{rgb}{0.21,0.49,0.74}
\usepackage[pagebackref,breaklinks,colorlinks,allcolors=wacvblue]{hyperref}

\def\wacvPaperID{196} 
\def\confName{WACV}
\def\confYear{2026}

\title{IPCD: Intrinsic Point-Cloud Decomposition}

\author{Shogo Sato, Takuhiro Kaneko, Shoichiro Takeda, Tomoyasu Shimada, \\Kazuhiko Murasaki, Taiga Yoshida, Ryuichi Tanida, Akisato Kimura\\
NTT Corporation\\
{\tt\small \{shg.sato, takuhiro.kaneko, shoichiro.takeda, tomoyasu.shimada,\}}\\
{\tt\small \{kazuhiko.murasaki, taiga.yoshida, ryuichi.tanida, akisato.kimura\}@ntt.com}
}

\begin{document}
\maketitle
\begin{abstract}
Point clouds are widely used in various fields, including augmented reality (AR) and robotics, where relighting and texture editing are crucial for realistic visualization. Achieving these tasks requires accurately separating albedo from shade. However, performing this separation on point clouds presents two key challenges: (1) the non-grid structure of point clouds makes conventional image-based decomposition models ineffective, and (2) point-cloud models designed for other tasks do not explicitly consider global-light direction, resulting in inaccurate shade. In this paper, we introduce \textbf{Intrinsic Point-Cloud Decomposition (IPCD)}, which extends image decomposition to the direct decomposition of colored point clouds into albedo and shade. To overcome challenge (1), we propose \textbf{IPCD-Net} that extends image-based model with point-wise feature aggregation for non-grid data processing. For challenge (2), we introduce \textbf{Projection-based Luminance Distribution (PLD)} with a hierarchical feature refinement, capturing global-light ques via multi-view projection. For comprehensive evaluation, we create a synthetic outdoor-scene dataset. Experimental results demonstrate that IPCD-Net reduces cast shadows in albedo and enhances color accuracy in shade. Furthermore, we showcase its applications in texture editing, relighting, and point-cloud registration under varying illumination. Finally, we verify the real-world applicability of IPCD-Net.
\end{abstract}

\begin{figure}
\centering
\includegraphics[width=0.85\linewidth]{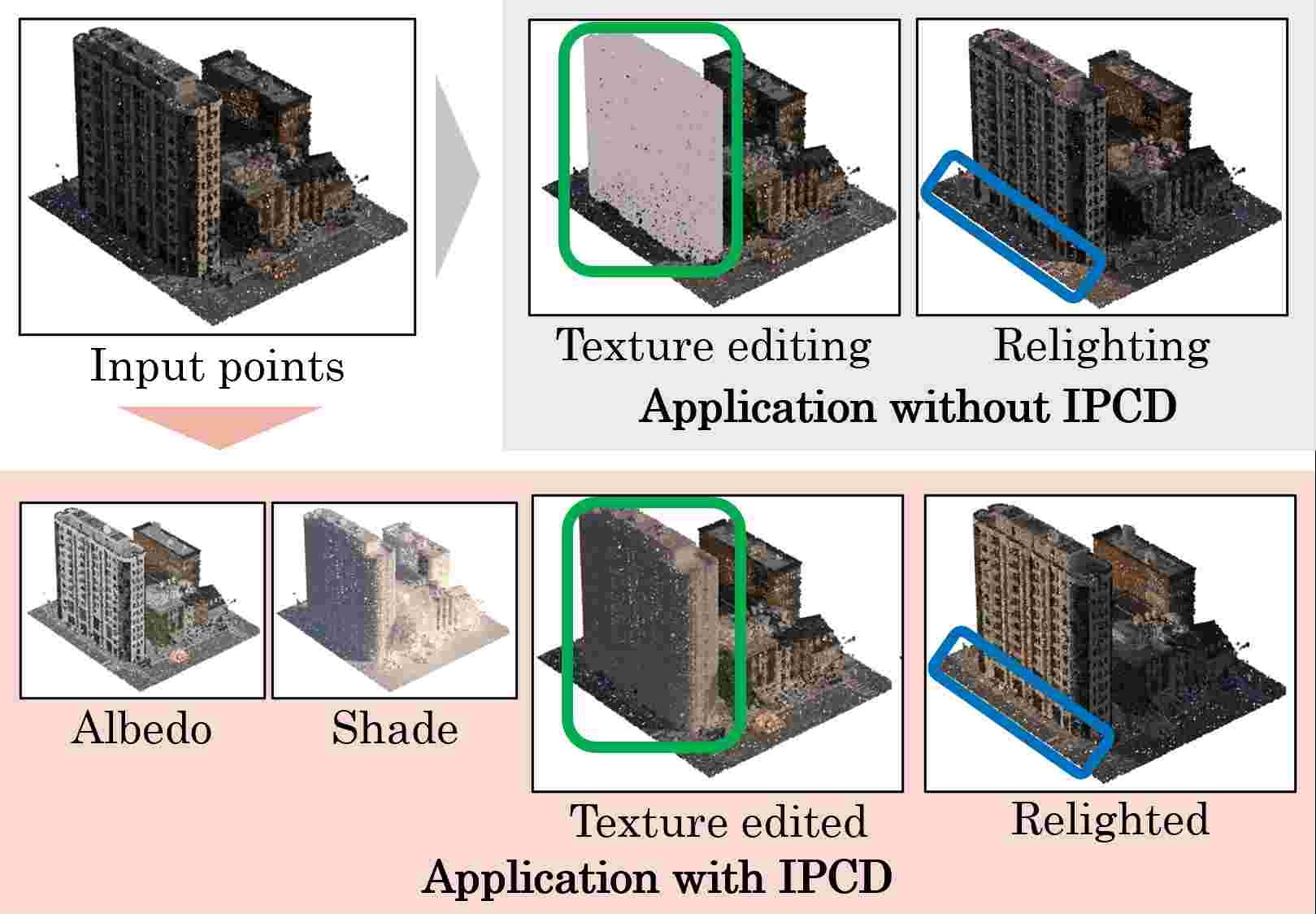}
\vspace{-3mm}
\caption{\label{fig1} 
Motivation of Intrinsic Point-Cloud Decomposition (IPCD). (a) Directly using colored point clouds for texture editing and relighting introduces artifacts due to illumination, resulting in unnatural colors (green) and residual shade (blue). To address this issue, (b) we extend image decomposition to point clouds, namely \textbf{IPCD}, which decomposes point clouds into albedo and shade. Applying IPCD enables more natural texture editing and significantly reduces residual shade in relighting.}
\vspace{-5mm}
\end{figure}

\vspace{-3mm}
\addtocontents{toc}{\protect\setcounter{tocdepth}{-1}}
\section{Introduction\label{sec:intro}}
Colored point clouds are widely used in applications such as augmented reality (AR)~\cite{mahmood2020bim} and robotics~\cite{kim2018slam}, due to their ability to preserve geometry and appearance~\cite{guo2020deep}. These applications often require texture editing and relighting to enhance visual realism and simulate diverse illuminations~\cite{sabbadin2019high, gao2024relightable}. Additionally, point-cloud registration is required for reconstructing digital spaces even under varying illuminations~\cite{ren2021color, colorpcr}. However, the entanglement of surface reflectance (albedo) and illumination (shade) in colored point clouds may hinder these tasks as shown in~\cref{fig1}. With the widespread of three dimensional (3D) scanners equipped with synchronized cameras such as NavVis~\cite{navviswebsite}, only colored point clouds are often stored, while the original images are discarded. Thus, techniques for directly decomposing colored point clouds are desired.

In two-dimensional (2D) image domain, intrinsic image decomposition (IID)~\cite{land1971, barrow1978} has been extensively explored to decompose an image into albedo and shade~\cite{careaga2024colorful, chen2025intrinsicanything}. To apply existing IID models to point cloud, we render the points into images for IID and project the results back to points, but this may lead to low decomposition quality due to occlusions and geometric lacking. 

Achieving direct point-cloud decomposition introduces two key challenges: (1) the non-grid structure of point clouds makes conventional image-based decomposition models ineffective, and (2) point-cloud models designed for other tasks do not explicitly consider global-light direction, resulting in inaccurate shade.

In this paper, we extend the conventional IID problem to colored point clouds, which we refer to as \textbf{Intrinsic Point-Cloud Decomposition (IPCD)}. IPCD decomposes a colored point cloud into albedo and shade\footnote{Note that, global-light direction is not provided as input, aligning with real-world scenarios where such information is typically unavailable.}. To tackle IPCD, we propose \textbf{IPCD‑Net}, a deep learning model designed to address the aforementioned challenges (1) and (2). First, to address challenge (1), we extend IID models with point-wise feature aggregation based on the point-cloud model for other tasks~\cite{wu2022point}, enabling effective processing of non-grid data. Specifically, we propose IPCD-Net$_{\text{Base}}$, inspired by PoInt-Net~\cite{xing2023intrinsic}, which performs IID on RGB-D data via point-cloud representations but computes losses in image space. In contrast, IPCD-Net$_{\text{Base}}$ performs IPCD entirely in point-cloud space, allowing direct optimization in the 3D geometric domain. Second, to address challenge (2), we introduce \textbf{Projection-based Luminance Distribution (PLD)}. PLD captures global-light cues by projecting the input point cloud from multiple viewpoints on a hemispherical surface and adopting its average luminance as a feature representation, as detailed in~\cref{fig3}. To effectively integrate PLD features, we employ a hierarchical feature refinement that progressively estimates albedo and shade. Then, we propose \textbf{IPCD-Net}, which combines IPCD-Net$_{\text{Base}}$ and PLD with hierarchical feature refinement to achieve robust IPCD performance.

Additionally, we create a synthetic dataset of colored point clouds with albedo and shade ground truths in outdoor scenes to assess IPCD performance. Based on this dataset, we verify the effectiveness of IPCD-Net, practicality in reducing cast shadows and improving shade color accuracy. Additionally, we verify the practicality of IPCD-Net in texture editing, relighting, and point-cloud registration under varying illumination. Finally, we verify the real-world applicability of IPCD-Net by applying it to real-world colored point clouds captured by a 3D scanner. 

The main contributions are summarized as follows.
\begin{itemize}
  \item We extend IID to colored point clouds and formalize this setting as \textbf{Intrinsic Point-Cloud Decomposition (IPCD)}, decomposing a colored point cloud into albedo and shade. To assess IPCD performance, we create an synthetic dataset of colored point clouds with albedo and shade ground truths.
  \item  We propose \textbf{IPCD-Net}, an IPCD-specific model that incorporates point-wise feature aggregation to handle non-grid data. IPCD-Net also employs \textbf{Projection-based Luminance Distribution (PLD)} with a hierarchical feature refinement to capture global-light cues.
  \item We verify the practicality of IPCD-Net in texture editing, relighting, and registration under varying illumination. Additionally, we verify its real-world applicability.
\end{itemize}

\section{Related works\label{sec2}}
This section summarizes point-cloud models, IID models, and light-direction estimation, which form the foundation of our IPCD approach.\smallskip\\
\textbf{Point-cloud models.}
Point-cloud models are widely explored in classification~\cite{hackel2017semantic3d, uy2019revisiting} and semantic segmentation~\cite{wang2019graph, zhang2020polarnet} tasks. Early models employed shared-weight multilayer perceptrons (MLPs) to extract point-wise features~\cite{qi2017pointnet}. Later, multi-scale grouping was introduced~\cite{qi2017pointnet++} and its further refined version by subsequent architectures~\cite{ni2020pointnet++, qian2022pointnext}. To better handle the non-grid structure of point clouds, PointCNN~\cite{li2018pointcnn} introduced a grid-like convolutional approach. More recently, Point Transformer (PTv1)~\cite{zhao2021point} leverages self-attention mechanisms~\cite{vaswani2017attention} for enhanced feature interactions. Point Transformer version 2 (PTv2)~\cite{wu2022point} further optimized this by grouped vector attention. The recent model~\cite{wu2024ptv3}, enhances computational efficiency using serialized neighbor mapping. The aforementioned models assign per-point labels from colored point clouds, making them as baselines for IPCD, which requires estimating per-point albedo and shade.\smallskip\\
\textbf{Intrinsic image decomposition (IID).}
IID decomposes an image into albedo and shade. Early methods estimated only albedo, and then computed shade via image division~\cite{zhou2015, fan2018}. Later approaches simultaneously estimated albedo and shade~\cite{li2018}, by employing shared feature extractors with respective decoders for both~\cite{liu2017unsupervised, zhou2019glosh, seo2021}. Recent work, PoInt-Net~\cite{xing2023intrinsic}, performs IID using RGB-D data by leveraging geometric information through point-cloud representation. However, PoInt-Net is designed for image-based IID, not for point-cloud decomposition. Beyond single-image IID, multi-view methods, such as Neural Radiance Fields (NeRF)~\cite{mildenhall2021nerf} and 3D Gaussian Splatting~\cite{kerbl20233d}, have been explored for IID~\cite{zhang2021physg, zhu2023i2, liang2024gs}. However, their direct application to IPCD is challenging due to occlusions in point-cloud rendering and per-scene training limitations, which hinder generalization across diverse scenes.\smallskip\\
\textbf{Light-direct estimation.}
The global-light feature is critical cue for accurate albedo and shade decomposition in IPCD. Traditional methods estimate light direction from a single image using shade cues~\cite{barron2015shape} or weak signals from environmental elements~\cite{zhang2020light}. Recent works explore the task from multiple images~\cite{choi2023mair, zhao2023multi}, leveraging global-light consistency. However, these approaches are designed for image-based estimation and are not directly applicable to point clouds. To address this, we propose Projection-based Luminance Distribution (PLD), which captures the correlation between projected image brightness and global-light direction.

\begin{figure*}
\centering
\includegraphics[width=1.0\linewidth]{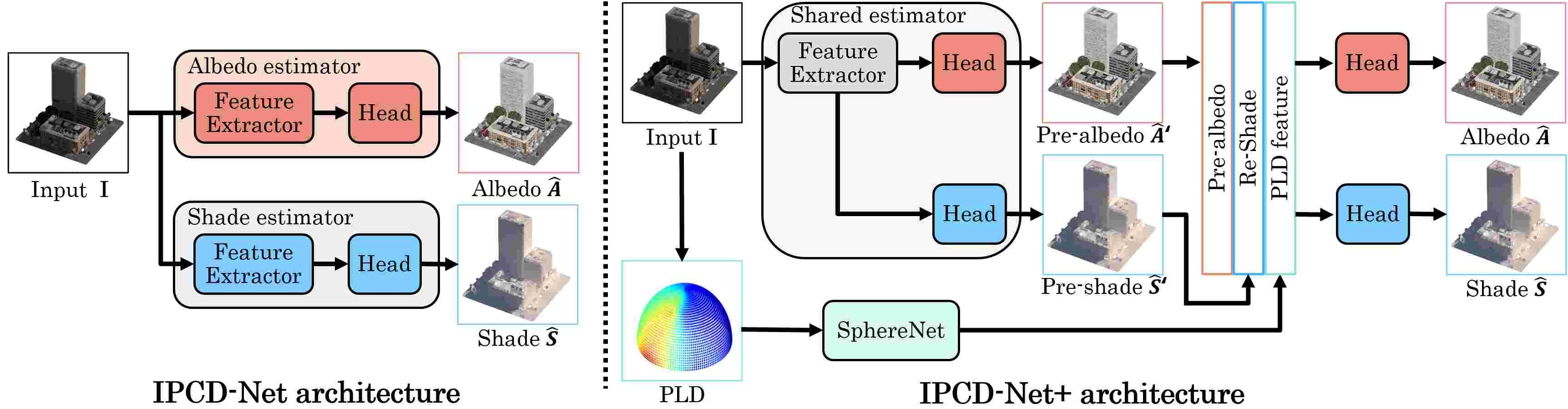}
\vspace{-5mm}
\caption{\label{fig2} 
Architectures of (a) IPCD-Net$_{\text{Base}}$ and (b) IPCD-Net. IPCD-Net$_{\text{Base}}$ independently estimates albedo $\hat{A}$ and shade $\hat{S}$ from the colored point cloud $I$ using separate albedo and shade estimators. IPCD-Net employs a shared feature extractor to process $I$, generating pre-albedo and pre-shade representations ($\hat{A'}, \hat{S'}$). The Projection-based Luminance Distribution (PLD) is processed by SphereNet to capture global-light features and enhance IPCD performance. The final albedo $\hat{A}$ and shade $\hat{S}$ are obtained by integrating PLD features with the pre-albedo and pre-shade representations through a hierarchical feature refinement.}
\vspace{-3mm}
\end{figure*} 
\section{Methods\label{sec3}}
This section begins with the problem setting for IPCD. We then describe IPCD-Net$_{\text{Base}}$, an expanded version of PoInt-Net~\cite{xing2023intrinsic} for point cloud decomposition. Finally, we introduce IPCD-Net, which incorporates PLD with a hierarchical feature refinement to enhance decomposition accuracy.

\subsection{Problem setting}
The IPCD task aims to estimate albedo $\hat{A}\in\mathbb{R}^{N\times3}$ and shade $\hat{S}\in\mathbb{R}^{N\times3}$ from a colored point cloud defined by color $I\in\mathbb{R}^{N\times3}$ and position $P\in\mathbb{R}^{N\times3}$, where $N$ represents the number of the points. Each point $i$ consists of 3D position $P_i$ = ($x_i, y_i, z_i$) and color $I_i$ = ($I_i^r, I_i^g, I_i^b$). The goal is to estimate albedo color $\hat{A}_i = (\hat{A}_i^r, \hat{A}_i^g, \hat{A}_i^b)$ and shade color $\hat{S}_i = (\hat{S}_i^r, \hat{S}_i^g, \hat{S}_i^b)$. During training, the model takes input color $I$ and position $P$, learning to estimate albedo $\hat{A}$ and shade $\hat{S}$ using ground-truth albedo $A\in\mathbb{R}^{N\times3}$ and shade $S\in\mathbb{R}^{N\times3}$. Unlike NeRF-based approaches that require per-scene optimization, IPCD-Net can perform IPCD on arbitrary point clouds without requiring scene-specific fine-tuning. During inference, the model receives $I$ and  $P$, and estimates $\hat{A}$ and $\hat{S}$. To verify the IPCD performance, we prepare IPCD dataset, containing multiple scene of point cloud pairs ($P, I, A, S$), as described in~\cref{sec6-1+}. Note that global-light direction and illumination color are not provided as inputs, allowing IPCD models to be applied to real-world scenarios where such information is typically unavailable.

\subsection{Preliminary: PoInt-Net}
PoInt-Net~\cite{xing2023intrinsic} is an intrinsic image decomposition framework trained on RGB-D data ($I_{\rm{img}}, D_{\rm{img}}$) paired with ground-truth light direction, shade image $S_{\rm{img}}$, and albedo image $A_{\rm{img}}$. This model consists of three modules; DirectionNet, Shader, and AlbedoNet. PoInt-Net first reconstructs colored point clouds from RGB-D input. Then, DirectionNet estimates the light direction, which is used alongside surface normals to compute rough shade. This rough shade is refined by Shader, producing the final shade points. AlbedoNet then estimates albedo points from the input point cloud. The estimated albedo and shade point clouds are projected back into image space ($\hat{A}_{\rm{img}}, \hat{S}_{\rm{img}}$). We then compute the albedo image loss $\mathcal{L}^{\rm{alb}}_{\rm{img}}$ and shade image loss $\mathcal{L}^{\rm{shd}}_{\rm{img}}$ by comparing ground-truth images. Given the assumption of a Lambertian surface, the product of albedo image $\hat{A}_{\rm{img}}$ and shade image $\hat{S}_{\rm{img}}$ is expected to match input image color $I_{\rm{img}}$~\cite{bell2014}. Thus, the physical loss $\mathcal{L}^{\rm{phy}}_{\rm{img}}$~\cite{liu2020} is also calculated. Additionally to these losses, PoInt-Net incorporates the gradient loss $\mathcal{L}^{\rm{gld}}_{\rm{img}}$ and the color cross ratio loss $\mathcal{L}^{\rm{ccr}}_{\rm{img}}$ to enforce smoothness and maintain color consistency. PoInt-Net follows a step-by-step learning strategy, where shade estimation modules are trained first, then albedo estimation is trained with shade parameters frozen. DirectionNet, Shader, and AlbedoNet employ MLP-based model~\cite{qi2017pointnet}.

\subsection{Expansion for IPCD: IPCD-Net\texorpdfstring{$_{\text{Base}}$}{Base}}
PoInt-Net~\cite{xing2023intrinsic} addresses intrinsic image decomposition using RGB-D input. This model computes supervision and loss in 2D image space. While this model shows its effectiveness in image-based settings, this model design makes it unsuitable for direct use in the IPCD task, which requires decomposition in unstructured 3D point-cloud space. To bridge this gap, we extend PoInt-Net and implement \textbf{IPCD-Net$_{\text{Base}}$}, which operates directly on colored point clouds with point-level supervision.
The network consists of two primary modules: Shade Estimator and Albedo Estimator as illustrated in~\cref{fig2}. The Shade Estimator predicts shade $\hat{S}$ directly from the input point cloud ($P, I$), bypassing the explicit light direction estimation in PoInt-Net due to the absence of ground-truth light direction. The Albedo Estimator then estimates albedo $\hat{A}$, following PoInt-Net’s AlbedoNet approach. A key difference is that PoInt-Net computes losses in image space, whereas IPCD-Net$_{\text{Base}}$ computes losses directly in point-cloud space as albedo loss $\mathcal{L}^{\rm{alb}}_{\rm{pnt}}$, shade loss $\mathcal{L}^{\rm{shd}}_{\rm{pnt}}$, and physical loss $\mathcal{L}^{\rm{phy}}_{\rm{pnt}}$.
\begin{equation}\label{eq4}
  \mathcal{L}^{\rm{alb}}_{\rm{pnt}}= \|A-\hat{A}\|_F,
\end{equation}
\begin{equation}\label{eq5}
  \mathcal{L}^{\rm{shd}}_{\rm{pnt}}= \|S-\hat{S}\|_F,
\end{equation}
\begin{equation}\label{eq6}
  \mathcal{L}^{\rm{phy}}_{\rm{pnt}}= \|I-\hat{A}\odot\hat{S}\|_F,
\end{equation}
where, $\|\cdot\|_F$  and $\odot$ represent the Frobenius norm and the Hadamard product, respectively. Due to the non-grid structure of point cloud, IPCD-Net$_{\text{Base}}$ does not utilize the gradient loss $\mathcal{L}^{\rm{gld}}$ or the color cross ratio loss $\mathcal{L}^{\rm{ccr}}$ employed in PoInt-Net. The training strategy follows PoInt-Net’s step-by-step learning scheme. IPCD-Net$_{\text{Base}}$ incorporates PTv2~\cite{wu2022point} instead of  MLP-based model~\cite{qi2017pointnet} for further effective feature extraction by grouped vector attention. Ptv2 is selected through preliminary validation.

\begin{figure}
\centering
\includegraphics[width=1.0\linewidth]{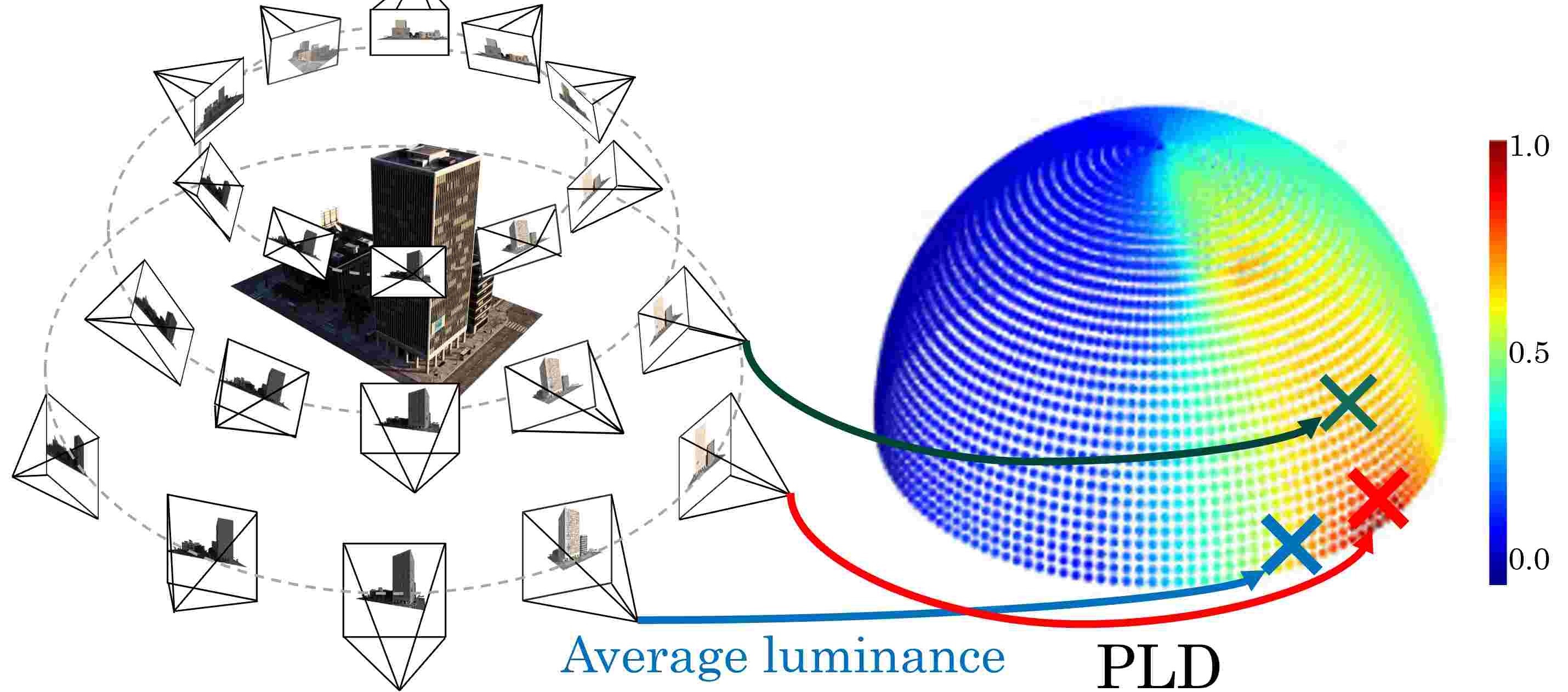}
\vspace{-5mm}
\caption{\label{fig3} PLD overview. The input point cloud is first projected onto an image plane from a specific viewpoint. By systematically varying the projection angles and computing the average luminance across these views, the PLD is constructed and represented on a hemispherical surface. Projections aligned with the illumination direction result in brighter images with minimal shade, whereas projections from the opposite direction yield darker images dominated by shadowed regions. The color bar represents the luminance intensity from its corresponding view.}
\vspace{-3mm}
\end{figure}

\subsection{Improvement for IPCD: IPCD-Net\label{sec4}}
While IPCD-Net$_{\text{Base}}$ enables direct decomposition on point clouds, it suffers from two issues: (i) color inconsistency in shade estimation and (ii) leakage between albedo and shade when trained independently. To address these, IPCD-Net introduces  a \textit{Projection-based Luminance Distribution (PLD)} to provide implicit global-light cues, and follows a \textit{simultaneous training approach} with a \textit{shared encoder}.\smallskip\\
\textbf{Projection-based luminance distribution (PLD).}
IPCD-Net$_{\text{Base}}$ often suffers from color inconsistency in shade estimation because it cannot account for global-light cues. In outdoor scenes, illumination color depends on the sun direction, but this information is unavailable in IPCD task setting. To address this, we design PLD to implicitly capture global-light cues by aggregating luminance from multiple viewpoints, as described in~\cref{fig3}. First, we render $256\times256$ images from $K=324$ uniformly sampled directions $(\theta,\phi)$ over the upper hemisphere (e.g., $\theta\in{0^{\circ},10^{\circ},\dots,80^{\circ}}, \phi\in{0^{\circ},10^{\circ},\dots,350^{\circ}}$), using point clouds with about $1.0\times10^{6}$ points within the field of view to ensure sufficient geometric fidelity. Each point cloud is rotated with the matrix $R(\theta,\phi)$, while the camera position $(0,0,1)$ and direction $(0,0,-1)$ are fixed. Next, we apply orthographic projection onto an image coordinates $(u, v)$ to generate the projected image $L(\theta, \phi; u, v)$. Subsequently, PLD value $\mathrm{PLD}(\theta, \phi)$ for direction $(\theta, \phi)$ is calculated as follows.
\begin{equation}\label{eq7}
\mathrm{PLD}(\theta, \phi) = \frac{1}{N_{\rm{P}}} \sum_{u,v} L(\theta, \phi; u, v),
\end{equation}
where $N_{\rm{P}}$ is the number of projected image pixels. Finally, we represent $\mathrm{PLD}(\theta, \phi)$ on a hemispherical surface. Projection in the illumination direction produces a brighter image with fewer shade, while projection in the opposite direction results in a darker image with dominant shade. Importantly, PLD does not assume a one-to-one mapping like ``bright to direct sunlight'' or ``dark to shade'', instead, it aggregates multi-view luminance distributions to derive statistical global-light cues. 
For PLD processing, standard CNNs are not suitable since they introduce significant distortions when projecting spherical data onto a 2D grid, especially near the poles. Thus, IPCD-Net employs SphereNet~\cite{coors2018spherenet}, directly handling spherical data by adjusting the sampling grid to the sphere’s geometry. SphereNet ensures distortion invariance and preserves the spatial relationships inherent to spherical representations.\smallskip\\
\textbf{Architecture.}
\cref{fig2} (b) illustrates the architecture of IPCD-Net, which improves over IPCD-Net$_{\text{Base}}$. IPCD-Net$_{\text{Base}}$ suffers from large model parameters and unstable training. To address this, IPCD-Net adopts a shared encoder that reduces parameters and stabilizes training by avoiding redundant learning of correlated albedo and shade features. Thus, first, the input point cloud ($P, I$) is processed by the shared encoder, followed by respective heads to estimate pre-albedo $\hat{A'}$ and pre-shade $\hat{S'}$. For natural embedding of global-light cues into both albedo and shade estimation, we employ a hierarchical feature refinement: intermediate predictions $\hat{A'}$ and $\hat{S'}$ are progressively refined with PLD features, allowing PLD to guide decomposition without overwhelming local cues. Second, the PLD feature $L\in\mathbb{R}^{3}$ is expanded to match the point cloud size $\mathbb{R}^{N\times3}$ and concatenated with $\hat{A'}$ and $\hat{S'}$. Finally, the concatenated features $\in\mathbb{R}^{N\times9}$ are processed by the albedo head and shade head to estimate final albedo $\hat{A}$ and shade $\hat{S}$, respectively.\smallskip\\
\textbf{Training strategy.}
IPCD-Net$_{\text{Base}}$ employs a step-by-step learning strategy, learning shade first, followed by albedo estimation. However, IPCD-Net follows a simultaneous training approach for both albedo and shade to capture their dependencies more effectively.\smallskip\\
\textbf{Losses.}
IPCD-Net estimates pre-albedo $\hat{A'}$, albedo $\hat{A}$, pre-shade $\hat{S'}$, and shade $\hat{S}$ from input point cloud $I$. We additionally define pre-albedo loss $\mathcal{L}^{\rm{alb}}_{\rm{pre}}$,  pre-shade loss $\mathcal{L}^{\rm{shd}}_{\rm{pre}}$, and pre-physical loss $\mathcal{L}^{\rm{phy}}_{\rm{pre}}$.
\begin{equation}\label{eq8}
  \mathcal{L}^{\rm{alb}}_{\rm{pre}}=\|A-\hat{A'}\|_F.
\end{equation}
\begin{equation}\label{eq9}
  \mathcal{L}^{\rm{shd}}_{\rm{pre}}=\|S-\hat{S'}\|_F.
\end{equation}
\begin{equation}\label{eq10}
  \mathcal{L}^{\rm{phy}}_{\rm{pre}}= \|I-\hat{A'}\odot\hat{S'}\|_F,
\end{equation}
Albedo loss $\mathcal{L}^{\rm{alb}}$, shade loss $\mathcal{L}^{\rm{shd}}$, and physical loss $\mathcal{L}^{\rm{phy}}$ are calculated like IPCD-Net$_{\text{Base}}$. In summary, the total loss $\mathcal{L}^{\rm{tot}}$ is defined as follows.

\begin{equation}
\mathcal{L}^{\rm{tot}}
 =\mathcal{L}^{\rm{alb}}_{\rm{pnt}}
 +\mathcal{L}^{\rm{shd}}_{\rm{pnt}}
 +\mathcal{L}^{\rm{phy}}_{\rm{pnt}}\\
 +\lambda \left( \mathcal{L}^{\rm{alb}}_{\rm{pre}}
 +\mathcal{L}^{\rm{shd}}_{\rm{pre}}
 +\mathcal{L}^{\rm{phy}}_{\rm{pre}} \right),
\end{equation}
where $\lambda$ is hyper parameters to balance these losses. Note that, all loss terms could be assigned independent hyper parameters to fine-tune their relative contributions. However, for simplicity and to avoid excessive hyper parameter tuning, we use a single $\lambda$ to collectively scale the pre-estimation losses.

\section{Experiments}\label{sec6}
This section starts with the experimental settings including compared models and implementation details in~\cref{sec6-1}. We then introduce IPCD dataset in \cref{sec6-1+} and evaluate IPCD performance on the dataset with an ablation study in \cref{sec6-2}. To demonstrate the practical applications of IPCD models, we showcase texture editing, relighting, and point-cloud registration under varying illumination conditions in \cref{sec6-3}. Finally, we apply IPCD models trained on synthetic data to real-world scenes for assessing generalization.

\subsection{Experimental setup\label{sec6-1}}
\textbf{Compared models.}
We compare IPCD-Net against both traditional baselines and recent IID models. First, we set Baseline-A ($\hat{A}=I, \hat{S}=1$) and Baseline-S ($\hat{S}=I, \hat{A}=1$) as the simplest baselines, treating the input point cloud $I$ directly as the estimated albedo or shade, following the typical IID evaluations~\cite{bell2014}. Retinex~\cite{grosse2009} is evaluated by extending it to IPCD. Recent IID models, including NIID-Net~\cite{luo2020}, CD‑IID\footnote{Due to its unavailable training code, this model was not fine tuned.\label{fn1}} \cite{careaga2024colorful}, IID‑Anything\footref{fn1}~\cite{chen2025intrinsicanything}, and GS‑IR~\cite{liang2024gs}, are applied to the IPCD dataset as references. Specifically, we render the point cloud into images, perform albedo and shade inference in 2D, then project the results back onto the point cloud. This pipeline may cause occlusion or artifacts, hence, it is not a strictly fair comparison to IPCD-Net, which operates directly on point clouds; however, we include these results as informative baselines.\smallskip\\
\textbf{Implementation details.}
All IPCD models are implemented using PyTorch~\cite{paszke2019pytorch} and trained on an NVIDIA H100 GPU. During training, 10,000 points are randomly sampled from each asset in every iteration to ensure diverse point coverage and address GPU memory limitations. Since the pre-estimation losses primarily serve as auxiliary supervision, we set $\lambda$ to a lower weight as $\lambda=0.1$.
The implementations of IPCD-Net are based on PTv2~\cite{wu2022point}\footnote{Github page: https://github.com/Pointcept/PointTransformerV2} and SphereNet~\cite{coors2018spherenet}\footnote{Github page: https://github.com/ChiWeiHsiao/SphereNet-pytorch}. PLD is computed with PyTorch3D renderer~\cite{ravi2020pytorch3d}, by varying both elevation and depression angles in 2.5 degree increments. The point size $\epsilon$ is set to 0.02. 

\tabcolsep = 2pt
\begin{table}
\centering
\small
\begin{tabular}{cccccccccc}
\toprule
\multirow{2}{*}{Model} & \multicolumn{2}{c}{MSE($10^{-2}$)$\downarrow$}& 
\multicolumn{1}{c}{}& \multicolumn{2}{c}{MAE($10^{-1}$)$\downarrow$}& 
\multicolumn{1}{c}{}& \multicolumn{2}{c}{PSNR$\uparrow$}\\
\cmidrule(lr){2-3}\cmidrule(lr){5-6}\cmidrule(lr){8-9}
&alb&shd&&alb&shd&&alb&shd\\
\midrule
Baseline-A~\cite{bell2014} &18.9&29.1&&3.58&4.27&&7.57&5.96\\
Baseline-S~\cite{bell2014} &30.4&25.1&&5.01&4.23&&5.22&6.39\\
Retinex~\cite{grosse2009}  &19.0&25.3&&3.59&4.28&&7.55&6.25\\
\hdashline
NIID-Net~\cite{luo2020} &15.2&12.1&&2.93&2.46&&8.97&9.99\\
CD-IID\footref{fn1}~\cite{careaga2024colorful} &17.3&27.8&&3.36&4.45&&8.00&5.88\\
IID-Anything\footref{fn1}~\cite{chen2025intrinsicanything} &16.5&27.1&&3.23&4.43&&8.35&6.04\\
GS-IR~\cite{liang2024gs}   &24.8&13.7&&4.41&2.78&&6.16&8.91\\
\hdashline
IPCD-Net$_{\text{Base}}$ &4.02&5.11&&1.58&1.62&&14.0&13.5\\
IPCD-Net &\textbf{3.03}&\textbf{3.25}
&&\textbf{1.31}&\textbf{1.37}&&\textbf{15.6}&\textbf{15.1}\\
\bottomrule
\end{tabular}
\caption{\label{table1} Numerical comparison of IPCD models. Recent IID models are limited in their performance due to dimensional reduction through rendering. IPCD models are trained in a supervised manner, leading to better performance than the rule-based and IID models. Especially, IPCD-Net demonstrated superior performance to IPCD-Net$_{\text{Base}}$ due to its incorporation of PLD, a shared feature extractor, and a hierarchical feature refinement.}
\vspace{-3mm}
\end{table}

\begin{figure*}
\centering
\includegraphics[width=0.86\linewidth]{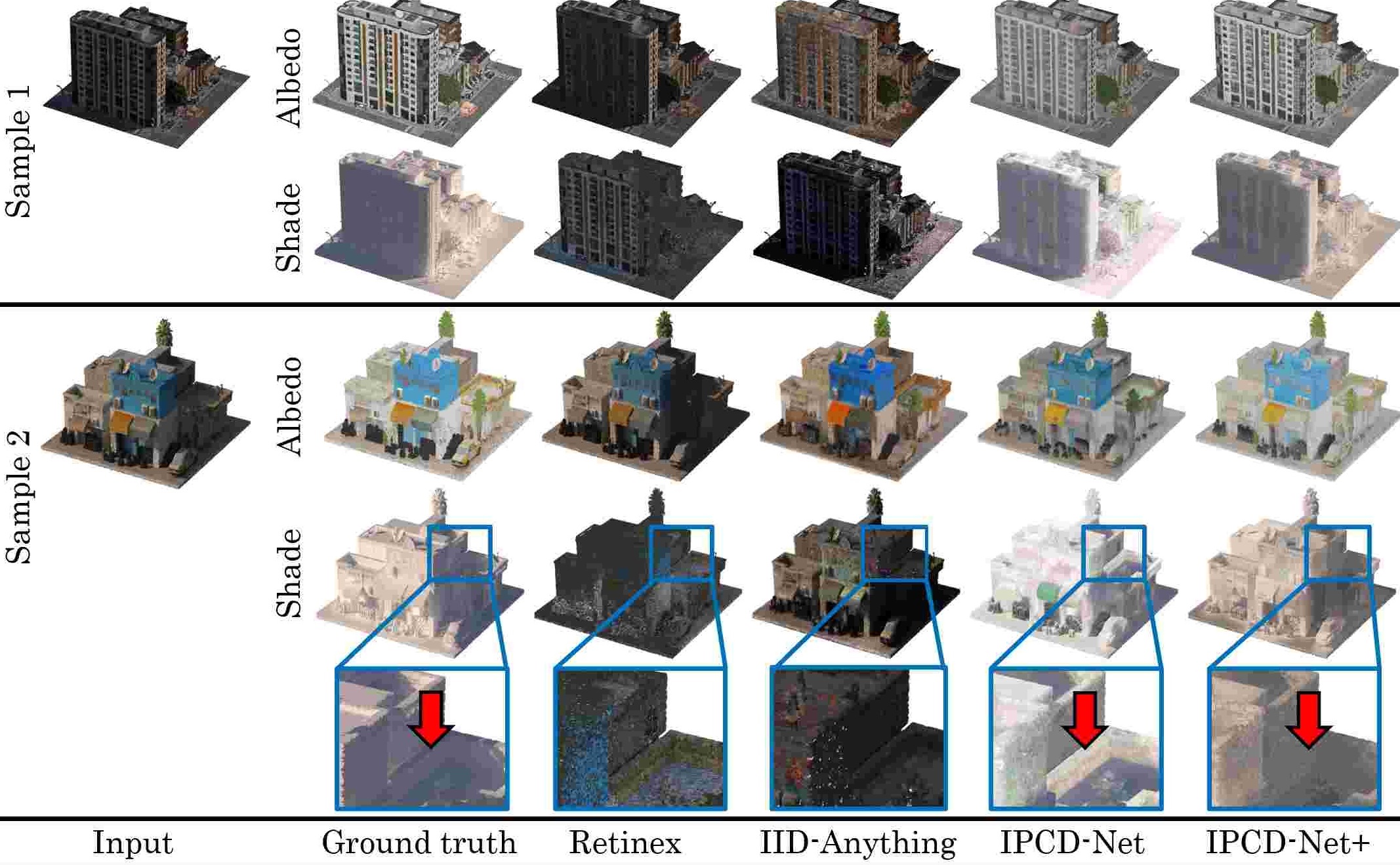}
\vspace{-3mm}
\caption{\label{fig5}Visual comparison of IPCD performance. Retinex~\cite{grosse2009} exhibits residual cast shadows, while more recent models mitigate this issue. IID-Anything~\cite{chen2025intrinsicanything} performs IPCD through point-cloud rendering, but inconsistencies in albedo brightness across rendered images introduce noise, and dimensional reduction during rendering leads to remaining shade. Our IPCD models effectively separate albedo from shade. However, IPCD-Net$_{\text{Base}}$ lacks explicit light direction, causing some cast shadows to be omitted in shade, as indicated by red arrows. Conversely, IPCD-Net successfully estimates the shade color and cast shadows.}
\vspace{-3mm}
\end{figure*}

\subsection{Dataset preparation\label{sec6-1+}}
To train and evaluate IPCD models, we introduce the IPCD dataset, a synthetic 3D dataset specifically designed for IPCD. We focused on the most common and practical case: outdoor clear-sky scenes with a few buildings of diffuse materials, illuminated by a sunlit directional light. The dataset construction process consists of the following steps:
\begin{enumerate}
  \item Prepare surface meshes of synthetic outdoor scenes with albedo colors, and label them as ``albedo''.
  \item Simulate sunlight based on geographic location and three different time of day to illuminate the albedo meshes, labeling the results as ``input''.
  \item Remove albedo colors from the meshes and re-illuminate the meshes under identical conditions to obtain ``shade''.
  \item Export the meshes as point clouds by random sampling with 1,000,000 points on the mesh surface.
\end{enumerate}
Consequently, the dataset includes 90 point-cloud sets derived from 30 unique assets. To prevent data leakage and ensure model generalizability, we divide the dataset into training and test sets, with 23 and 7 assets respectively. Supplementary material provides more details on the dataset construction process.

\subsection{Results and discussion\label{sec6-2}}
\textbf{Quantitative results.}
\cref{table1} summarizes the performance of IPCD-Net and the compared models. Baseline-A and Baseline-S serve as references, directly using the input point cloud as albedo and shade, respectively. Since Retinex primarily aims to reduce attached shadows by smoothing regions with similar neighboring colors, it is less effective on the IPCD dataset, where cast shadows dominate. Recent IID models, including CD-IID, IID-Anything, and GS-IR, apply IPCD through point-cloud rendering. Their performance is limited due to dimensional reduction during rendering, which loses geometric details crucial for intrinsic decomposition. In contrast, IPCD-Net$_{\text{Base}}$ estimates both albedo and shade directly in point-cloud space, achieves superior quantitative performance. IPCD-Net further enhances results by integrating PLD, a shared feature extractor, and a hierarchical feature refinement. The contributions of each component in IPCD-Net are analyzed in the ablation study.\smallskip\\
\textbf{Visual results.}
\cref{fig5} provides qualitative comparisons of albedo and shade estimations by IPCD-Net and the compared models, complementing the quantitative results in \cref{table1}. For clarity, we showcase representative examples from each category; additional results are provided in the supplementary materials. Retinex smooths regions of similar colors, producing albedo estimates that closely resemble the input point cloud, failing to remove cast shadows. Recent IID models exhibit inconsistencies in albedo brightness across rendered images, introducing noise, and dimensional reduction during rendering leads to residual shadows.
IPCD-Net$_{\text{Base}}$ provides visually improved results due to its direct estimation in point-cloud space. However, without explicit global light direction input, IPCD-Net$_{\text{Base}}$ may omit some cast shadows in shade estimation as indicated by red arrows. Additionally, the lack of light direction input causes the shade color inaccuracy. In contrast, IPCD-Net maintains the original colors while effectively separating albedo from shade, demonstrating its robustness in IPCD. Additional examples are provided in Supplementary material.\smallskip

\tabcolsep = 2pt
\begin{table}
\centering
\small
\begin{tabular}{cccccccccc}
\toprule
\multirow{2}{*}{Model} & \multicolumn{2}{c}{MSE($10^{-2}$)$\downarrow$}& 
\multicolumn{1}{c}{}& \multicolumn{2}{c}{MAE($10^{-1}$)$\downarrow$}& 
\multicolumn{1}{c}{}& \multicolumn{2}{c}{PSNR$\uparrow$}\\
\cmidrule(lr){2-3}\cmidrule(lr){5-6}\cmidrule(lr){8-9}
&alb&shd&&alb&shd&&alb&shd\\
\midrule
w/o PLD         &3.00&3.46&&1.30&1.40&&15.6&14.9\\
w/o HFR+PLD     &3.24&3.41&&1.35&1.38&&15.5&14.9\\
w/o share enc.  &3.16&3.32&&1.33&1.33&&15.4&15.2\\
\midrule
IPCD-Net        &3.03&3.25&&1.31&1.37&&15.6&15.1\\
\bottomrule
\end{tabular}
\vspace{-3mm}
\caption{\label{table2} Ablation study to evaluate the contribution of each component in IPCD-Net, including PLD, a shared encoder, and a hierarchical feature refinement (HFR).}
\vspace{-3mm}
\end{table}

\noindent
\textbf{Ablation study.}
To evaluate the contribution of each component in IPCD-Net, we conducted an ablation study by selectively removing or modifying specific elements. \cref{table2} presents the quantitative impact of PLD, a shared encoder, and a hierarchical feature refinement. PLD contributes to shade estimation performance, especially for shade colors, since the illumination color of outdoor scenes varies with the angle of the sun. The hierarchical feature refinement with pre-albedo and pre-shade concatenation supports albedo estimation from shade priors, hence, "w/o PLD" represents better albedo estimation than "w/o HFR+PLD". The shared-feature extractor captures a common feature representation for both albedo and shade to enhance these estimation performance. Further detailed ablation study is provided in Supplementary material.

\subsection{Applications\label{sec6-3}}

\begin{figure}
\centering
\includegraphics[width=0.9\linewidth]{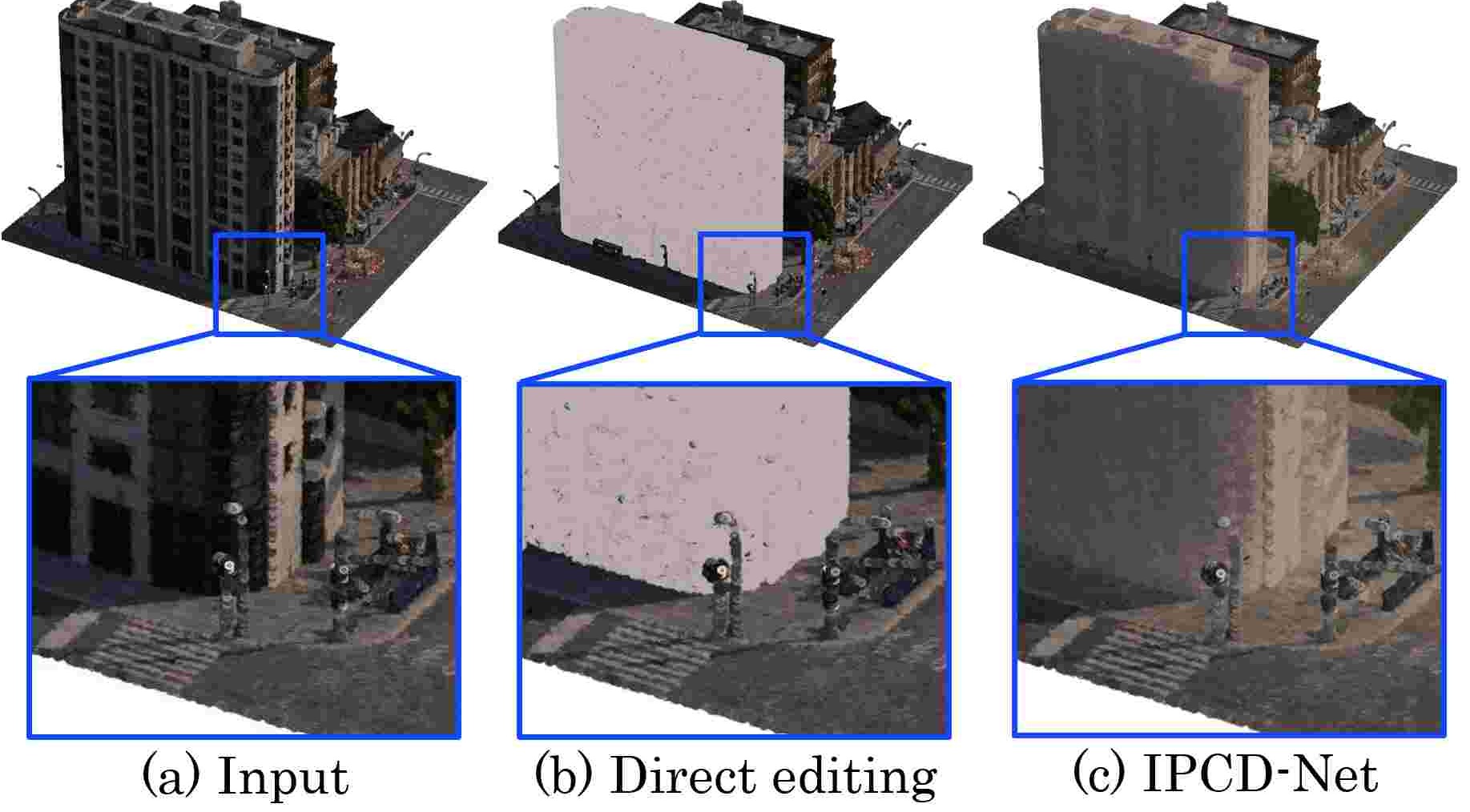}
\vspace{-3mm}
\caption{\label{fig7} Visual result of texture editing with IPCD. (a) Input point cloud, (b) direct editing, and (c) editing through IPCD-Net are shown. Direct texture editing results in unnatural appearance due to lighting inconsistencies. Editing albedo separated by IPCD-Net, then recombining with shade, achieving natural results.}
\vspace{-4mm}
\end{figure}

\noindent
\textbf{Texture editing.}  
Directly modifying point cloud colors often results in unnatural appearances due to the presence of embedded shade effects. To overcome this limitation, IPCD-Net decomposes point clouds into albedo and shade components, enabling albedo editing while preserving the original shade. This decomposition facilitates more natural texture modifications by ensuring that shade remains consistent even after color adjustments. To assess the effectiveness of this approach, we conducted a targeted color adjustment on building structures by altering specific regions of the input illumination $I$ to a gray tone, referred to as direct editing. Additionally, we modified the underlying albedo $\hat{A}$ while maintaining the shade $\hat{S}$, following IPCD-based editing. \cref{fig7} compares direct color editing with IPCD-based editing, demonstrating that direct modifications introduce visual inconsistencies due to unintended shade alterations. In contrast, IPCD-based editing produces realistic textures. Unlike NeRF-based methods~\cite{liang2024gs}, which require per-scene optimization, IPCD-Net can perform texture editing across diverse point cloud datasets without requiring scene-specific fine-tuning. \smallskip

\begin{figure}
\centering
\includegraphics[width=1.0\linewidth]{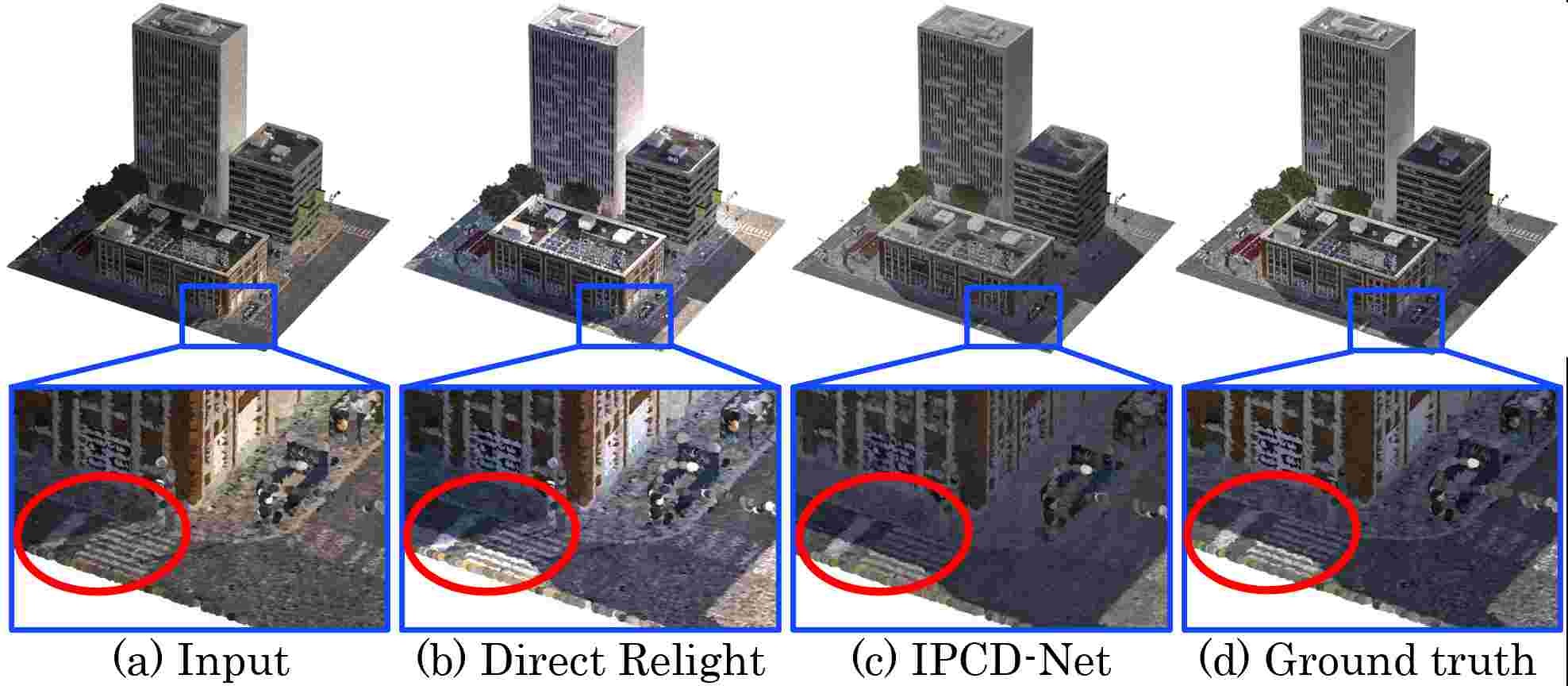}
\vspace{-5mm}
\caption{\label{fig8} Visual results of relighting application, including (a) input, (b) direct relighting, (c) relighting through IPCD-Net, and (d) ground truth. Direct relighting led to remaining cast shadow of the original points, as indicated by red circle. IPCD models reduce cast shadows and relight points similar to ground truth.}
\vspace{-5mm}
\end{figure}

\noindent
\textbf{Relighting.}  
Altering illumination in colored point clouds without intrinsic decomposition often result in unrealistic appearances due to the entanglement of albedo and shade. The IPCD framework addresses this issue by decomposing albedo from shade, allowing for the simulation of diverse lighting conditions while preserving the intrinsic color properties of objects. To evaluate the effectiveness of this approach, we utilized the IPCD dataset, which provides albedo under three distinct lighting conditions. Specifically, we selected an input point cloud \(I_1\) consisting of albedo \(A_1\) and shade \(S_1\) and replaced the illumination condition from \(S_1\) to \(S_2\) for relighting. Direct relighting is performed by directly combining \(I_1\) with \(S_2\) due to the inherent challenges in separating albedo and shade. Instead, IPCD-Net first decomposes \(I_1\) into \(\hat{A}_1\) and \(\hat{S}_1\), after which \(\hat{A}_1\) is recombined with \(S_2\) to generate the relighted point cloud \(\hat{I}_{1\rightarrow2}\). As shown in \cref{fig8}, direct relighting results in residual cast shadows from the original shade, while IPCD-Net successfully reduces these artifacts and produces colors that closely resemble the ground truth. Notably, relighting through IPCD-Net also simulates accurate sunlit colors due to its precise decomposition. These findings demonstrate the effectiveness of IPCD in enabling realistic and flexible relighting of point clouds by separating albedo from shade.\smallskip

\begin{figure}
\centering
\includegraphics[width=0.9\linewidth]{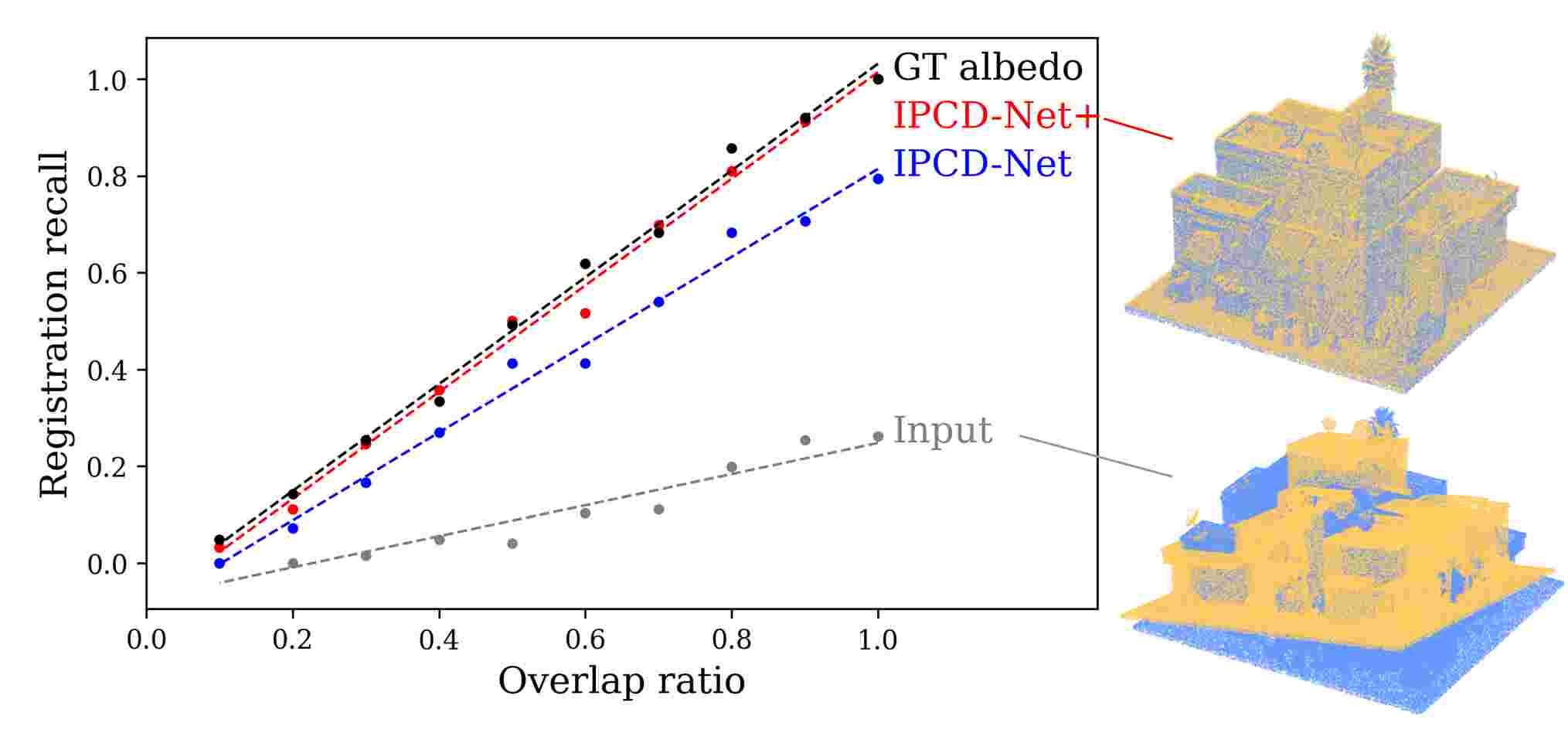}
\vspace{-3mm}
\caption{\label{fig6} Point-cloud registration under varying illumination. Registration performance is compared using raw input colors and estimated albedos, by registration recall (higher is better). The overlap ratio of input point clouds is varied to assess performance across different registration difficulties. Registration accuracy improves when the estimated albedo closely aligns with the ground truth, mitigating the effects of illumination variations. IPCD-Net achieves registration performance comparable to the ground truth, demonstrating its robustness under varying lighting conditions.}
\vspace{-4mm}
\end{figure}

\noindent
\textbf{Point-cloud registration.}
Aligning point clouds captured under varying lighting conditions presents significant challenges due to illumination-induced discrepancies. To address these inconsistencies, point-cloud registration should be performed using estimated albedo rather than directly relying on the original point clouds, which are affected by illumination variations. To evaluate the effectiveness of this approach, we conducted registration experiments on the IPCD dataset. Specifically, for each test asset, we obtained both the ground-truth albedo and the estimated albedo from IPCD models. Registration was then performed using pairs of albedo maps estimated from different sunlit times of the same asset, considering a total of 7 assets with 3 different time combinations. The alignment process was carried out using ColoredICP~\cite{park2017colored}, varying the overlap ratio, and the registration accuracy was evaluated using registration recall, as illustrated in~\cref{fig6}. When performing registration with the original input point clouds, variations in illumination conditions introduced inconsistencies that made the process more challenging. However, registration performance was significantly improved when using the albedo estimated by IPCD-Net$_{\text{Base}}$ and IPCD-Net. In particular, IPCD-Net achieved results comparable to the ground truth, demonstrating its robustness in aligning point clouds under varying illumination conditions.\smallskip

\begin{figure}
\centering
\includegraphics[width=0.85\linewidth]{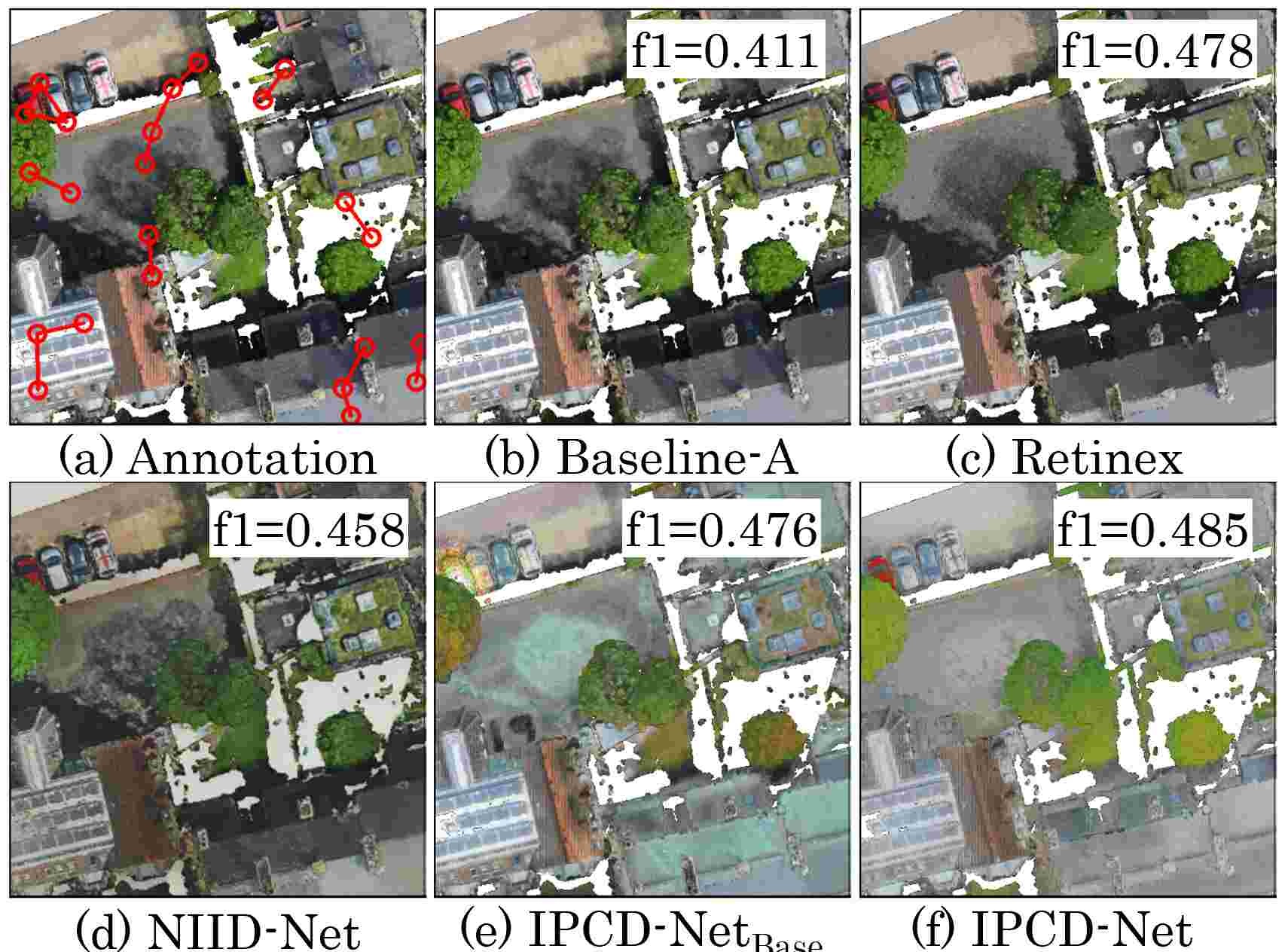}
\vspace{-3mm}
\caption{\label{fig9} Illustration of annotation and comparison results for real-world scenes of SensatUrban dataset~\cite{hu2022sensaturban}. (a) shows the annotation of relative reflectance between point pairs indicated by red circles. We also showcase (b) Baseline-A, (c) Retinex, (d) NIID-Net, (e) IPCD-Net$_{\text{Base}}$, and (f) IPCD-Net. We also show f1 score of each model for all 900 annotated point pairs. IPCD-Net achieves the best performance for visual and quantitative results, reducing cast shadows in real-world scenes.}
\vspace{-5mm}
\end{figure}

\noindent
\textbf{Real-world scenes.}
IPCD-Net can perform IPCD on real-world scenes without requiring scene-specific fine-tuning even when trained exclusively on synthetic data. To demonstrate its robustness, we evaluated IPCD-Net trained on the IPCD dataset using real-world point clouds from the SensatUrban dataset~\cite{hu2022sensaturban}, which consists of large-scale urban point clouds. Since the absolute values of ground truth albedo are difficult to obtain due to its inseparability from illumination, we conducted a quantitative evaluation with the relative reflectance between point pairs, following well-known method for real-world scenes~\cite{bell2014, sato2023}. Specifically, we rendered 12 SensatUrban samples and manually annotated 900 point-pairs with relative reflectance labels (darker/lighter/equal). We then evaluated IPCD models using the f1 score~\cite{sato2023}, showing that IPCD-Net out performs IPCD-Net$_{\text{Base}}$ and other models as depicted in \cref{fig9}. These results confirm IPCD-Net's robustness and applicability across diverse real-world environments. We further validated our approach on STPLS3D~\cite{chen2022STPLS3D}, Toronto-3D~\cite{tan2020toronto3d}, and KITTI~\cite{geiger2012we} datasets, observing similar trends as detailed in Supplementary material.

\subsection{IPCD-Net limitation}
IPCD-Net leverages PLD instead of explicit light direction, hence this approach has certain limitations. One challenge arises in scenes with highly heterogeneous reflectivity, such as a building with one highly reflective white surface and another with a low-reflective black surface. In such cases, the PLD may inaccurately estimate light direction due to the disproportionate influence of surfaces with strong reflectivity differences. However, these extreme cases are uncommon and typically confined to assets with atypical structures, limiting the practical impact of this issue. Another limitation pertains to scenarios involving sparse or occluded point clouds. When background elements are unexpectedly revealed, PLD may struggle to compute accurate luminance values due to missing information. In such cases, leveraging denser point clouds or integrating PLD with inpainting techniques~\cite{yu2018pu, qu2024conditional} could enhance model performance.

\section{Conclusion\label{sec7}}
We introduced IPCD, a task for decomposing colored point clouds into albedo and shade. To achieve this, we developed IPCD-Net, employing point-wise feature aggregation and PLD with a hierarchical feature refinement. IPCD-Net significantly outperformed IPCD-Net$_{\text{Base}}$, particularly in reducing cast shadows and improving shade color accuracy. It also demonstrated strong performance in texture editing, relighting, and point-cloud registration, with robustness validated on real-world scenes. As a first implementation, we adopted PTv2 as the feature extractor due to its effectiveness in point-cloud processing. However, IPCD-Net is adaptable to various point-cloud models, providing flexibility for further improvements. Future work will explore integrating advanced models~\cite{wu2024ptv3} to enhance its performance. 
{
    \small
    \bibliographystyle{ieeenat_fullname}
    \bibliography{main}
}

\end{document}